\documentclass[journal]{IEEEtran}
\usepackage{cite}
\usepackage{hyperref}
\usepackage{url}
\usepackage{helvet}
\usepackage{courier}
\usepackage{amssymb}
\usepackage{amsmath}
\usepackage{ragged2e}
\usepackage{times,amssymb,amsmath,amsthm,epsfig,graphicx,url}
\usepackage{pgfplotstable}
\usepackage{filecontents}
\usepackage{textcomp} 
\usepackage[english]{babel}
\usepackage{lipsum}
\usepackage{graphicx}
\usepackage{flushend}
\usepackage{stfloats}
\usepackage{xcolor}

\usepackage{epsfig}

\usepackage[switch,pagewise,columnwise]{lineno}
\usepackage[ruled,linesnumbered]{algorithm2e}
\usepackage[noend]{algpseudocode}
\usepackage{subfigure}
\usepackage{caption}
\usepackage{booktabs} % for professional tables
\usepackage{bm}
\usepackage{multirow}
\usepackage{threeparttable}
\usepackage{verbatim}
\usepackage{pifont}% http://ctan.org/pkg/pifont
\newcommand{\cmark}{\ding{51}}%
\newcommand{\xmark}{\ding{55}}%

\begin{document}
%\linenumbers
\title{Towards Accurate Knowledge Transfer via Target-awareness Representation Disentanglement}%

\author{Xingjian Li, Di Hu, Xuhong Li, Haoyi Xiong,~\IEEEmembership{Member, IEEE}, Chengzhong Xu,~\IEEEmembership{Fellow, IEEE}, Dejing Dou%\\
\IEEEcompsocitemizethanks{\IEEEcompsocthanksitem Xingjian Li, Xuhong Li, Haoyi Xiong, Dejing Dou are with Baidu Inc., Haidian, Beijing, 100085, China.\protect\\
\IEEEcompsocthanksitem Xingjian Li and Chengzhong Xu are with the State Key Lab of IOTSC, Faculty of Science and Technology, University of Macau, Macau SAR 999078, China.\protect\\
\IEEEcompsocthanksitem Di Hu is with the Gaoling School of Artificial Intelligence, Renmin University of China.
}
}


\markboth{Journal of \LaTeX\ Class Files,~Vol.~14, No.~8, August~2015}%
{Shell \MakeLowercase{\textit{et al.}}: Bare Demo of IEEEtran.cls for IEEE Journals}



\maketitle
\newcommand{\theHalgorithm}{\arabic{algorithm}}
\newcommand{\LTWO}{$\mathrm{L^2}$}
\newcommand{\LTWOSP}{$\mathrm{L^2\text{-}SP}$}
\newcommand{\DELTA}{$\mathrm{DELTA}$}
\newcommand{\BSS}{$\mathrm{BSS}$}
\newcommand{\AT}{$\mathrm{AT}$}
\newcommand{\TheName}{$\mathrm{TRED}$}

\begin{abstract}
Fine-tuning deep neural networks pre-trained on large scale datasets is one of the most practical transfer learning paradigm given limited quantity of training samples. To obtain better generalization, using the starting point as the reference (SPAR), either through weights or features, has been successfully applied to transfer learning as a regularizer. However, due to the domain discrepancy between the source and target task, there exists obvious risk of negative transfer in a straightforward manner of knowledge preserving. In this paper, we propose a novel transfer learning algorithm, introducing the idea of  \underline{T}arget-awareness  \underline{RE}presentation \underline{D}isentanglement (\TheName), where the relevant knowledge with respect to the target task is disentangled from the original source model and used as a regularizer during fine-tuning the target model. Specifically, we design two alternative methods, maximizing the Maximum Mean Discrepancy (Max-MMD) and minimizing the mutual information (Min-MI), for the representation disentanglement. Experiments on various real world datasets show that our method stably improves the standard fine-tuning by more than 2\% in average. \TheName\ also outperforms related state-of-the-art transfer learning regularizers such as \LTWOSP, \AT, \DELTA, and \BSS. 
\end{abstract}


% \fi
\IEEEpeerreviewmaketitle

\section{Introduction}
Deep convolutional networks achieve great success on large scale vision tasks such as ImageNet~\cite{russakovsky2015imagenet} and Places365~\cite{zhou2017places}. In addition to their notable improvements of accuracy, deep representations learned on modern CNNs are demonstrated transferable across relevant tasks~\cite{yosinski2014transferable}. This is rather fortunate for many real world applications with inefficient labeled examples. Transfer learning aims to obtain good performance on such tasks by leveraging knowledge learned by relevant large scale datasets. The auxiliary and desired tasks are called the source and target tasks respectively. According to~\cite{pan2009survey}, we focus on inductive transfer learning, which cares about the situation that the source and target tasks have different label spaces. 

The most popular practice is to fine-tune a model pre-trained on the source task with the \LTWO\ regularization, which has the effect of constraining the parameter around the origin of zeros.
\cite{li2018explicit} points out that since the parameter may be driven far from the starting point of the pre-trained model, a major disadvantage of naive fine-tuning is the risk of \textbf{catastrophic forgetting}~\cite{kirkpatrick2017overcoming} of the knowledge learned from source. They recommend to use the \LTWOSP\ regularizer instead of the popular \LTWO. While in parallel, knowledge distillation, which is originally designed for compressing the knowledge in a complex model to a simple one~\cite{hinton2015distilling}, is proved to be useful for transfer learning tasks where knowledge is distilled from a different dataset~\cite{zagoruyko2016paying,yim2017gift}. Recent work~\cite{li2019delta} formulates knowledge distillation in transfer learning as a regularizer on features and further improves through unactivated channel reusing for better fitting the training samples. These methods adopt a common assumption that, the relevant knowledge contained in the source model is expected to facilitate the generalization of the target model. This leads to the motivation of regularizing the fine-tuned model using the \emph{starting point as the reference} (SPAR) in a mainstream group of inductive transfer learning algorithms \cite{li2018explicit,zagoruyko2016paying,li2019delta}. 

%They propose improved regularization of \LTWOSP\ which instead encourages the parameter of the target model around the start point. Following the same idea of using the start point as reference,~\cite{li2019delta} argue that it is more reasonable to regularize middle representations of the network with attention rather than parameters. 

Although the advanced methods with SPAR have succeed to preserve the knowledge contained in the source model, fine-tuning also takes the obvious risk of  \textbf{negative transfer}~\cite{torrey2010transfer}.
%Apart from the aforementioned catastrophic forgetting, fine-tuning also takes the potential risk of  \textbf{negative transfer} for achieving effective knowledge learning from the source model.
Intuitively, if the source and target data distribution are dissimilar to some extent, not all the knowledge from the source is transferable to the target and an indiscriminate transfer could be detrimental to the target model. However, the impact of negative transfer has been rarely considered in inductive transfer learning studies. The most related work, \cite{chen2019catastrophic}, proposed to investigate the regularizer of Batch Spectral Shrinkage (\BSS) to inhibit negative transfer, where the small singular values of feature representations are considered as not transferable and suppressed. 
Yet, it is hard to adaptively determine the scope of small singular value when faced with different target tasks. Moreover, \BSS\ does not take consideration of the catastrophic forgetting risk, which means it has to be equipped with other fine-tuning techniques (e.g., \LTWOSP~\cite{li2018explicit}, Attention-Transfer~\cite{zagoruyko2016paying}, \DELTA~\cite{li2019delta}, etc.) to achieve considerable performance. %Table.~\ref{table:compare} shows a detailed comparison among these different fine-tuning techniques in terms of the catastrophic forgetting and negative transfer risk.

According to the above analysis, it is straightforward to think about a better solution which simultaneously takes the consideration of preserving relevant knowledge and avoiding negative transfer. 
In this paper, we intend to improve the standard fine-tuning paradigm by accurate knowledge transfer. Assuming that the knowledge contained in the source model consists of one part relevant to the target task and the other part which is irrelevant\footnote{Note that although this won't be mathematically guaranteed, it is very common in practice that the source task is for general purpose while the target task focuses on a specific domain.}, we are going to explicitly disentangle the former from the source model. Thus, a target task specific starting point is used as the reference instead of the original one. 
Specifically, we design a novel regularizer of deep transfer learning through \underline{T}arget-awareness  \underline{RE}presentation \underline{D}isentanglement (\TheName). The whole algorithm includes two steps. First we use a lightweight disentangler to separate middle representations of the pre-trained source model into the positive and negative parts. The core component of the disentangler is a differentiable module which is capable of separating the positive and negative group of features. We propose two alternative methods for the disentanglement, which are maximizing the Maximum Mean Discrepancy (Max-MMD) on the spatial dimension, and minimizing the Mutual Information (Min-MI) on the channel dimension. Intuitively, we intend to distill the relevant ingredients from noisy features by separating irrelevant ingredients which should: (1) pay attention to different regions with the the relevant ones; or (2) have independent semantic representations with the relevant ones. The disentanglement is achieved by simultaneously optimizing the disentangling module and ensuring to reconstruct the original representation. Supervision information from labeled target examples is utilized to distinguish the positive part from the negative part. The second step is to perform fine-tuning using the disentangled positive part of representations as the reference. In summary, our main contributions are as following: 
\begin{itemize}
\item We are the first to involve the idea of representation disentanglement to improve inductive transfer learning.  
\item Our algorithm aiming at accurate knowledge transfer contributes to the study of negative transfer.
\item Our proposed \TheName\ significantly outperforms state-of-the-art transfer learning regularizers including \LTWO, \LTWOSP, \AT, \DELTA\ and \BSS\ on various real world datasets. 
\end{itemize}

\begin{threeparttable}[t]
\centering
\caption{Comparison among different fine-tuning approaches.}
\label{table:compare}
\begin{tabular}{p{3cm}|p{1.5cm}p{1.5cm}}
\multirow{2}*{Approach}& \multicolumn{2}{c}{Risk $Considered$} \\
                      & CF\tnote{*} & NT\tnote{*} \\\hline
\LTWO~\cite{donahue2014decaf}  & \ \xmark & \ \xmark \\
SPAR~\cite{zagoruyko2016paying,li2018explicit,li2019delta}  &  \ \cmark & \ \xmark \\
BSS~\cite{chen2019catastrophic}     & \ \xmark & \ \cmark  \\
\textbf{TRED} &  \ \cmark & \ \cmark
\end{tabular}
\begin{tablenotes}
 \footnotesize
 \item[*] CF=Catastrophic Forgetting, NT=Negative Transfer, SPAR=Starting Point As the Reference.
\end{tablenotes}
\end{threeparttable}

\begin{figure*}[t]
\begin{center}
\includegraphics[height=0.43\textwidth,width=0.9\textwidth]{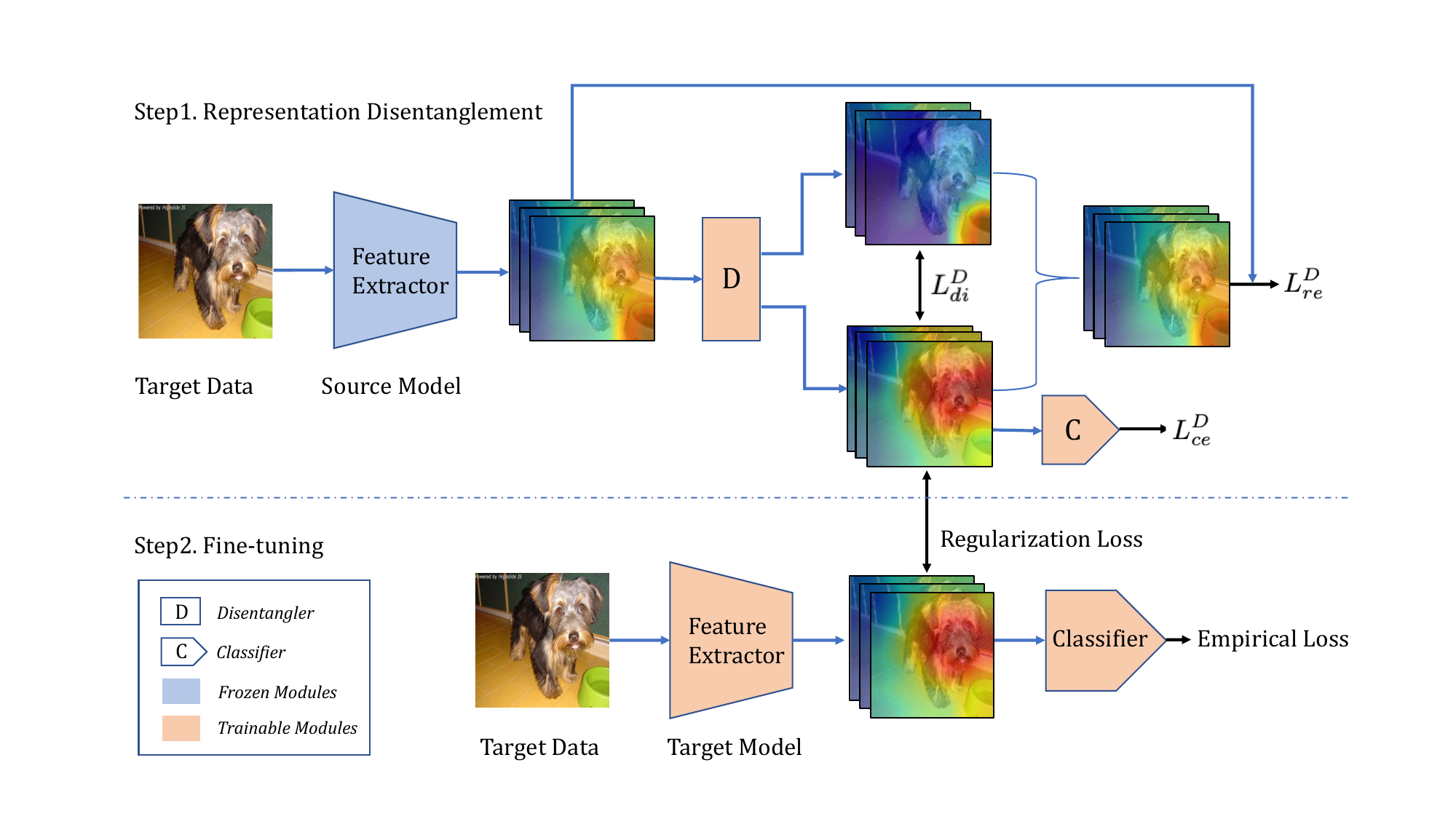}
\caption{The architecture of deep transfer learning through Target-awareness Representation Disentanglement (\TheName).}
\label{fig:arch}
\end{center}
\end{figure*}

\section{Related Work}
\subsection{Shrinkage Towards Chosen Parameters}
Regularization techniques have a long history since Stein's paradox~\cite{stein1956inadmissibility,efron1977stein}, showing that shrinking towards chosen parameters obtains an estimate more accurate than simply using observed averages. Most common choices like Lasso and Ridge Regression pull the model towards zero, while it is widely believed that shrinking towards ``true parameters" is more effective. In transfer learning, models pre-trained on relevant source tasks with sufficient labeled examples are often regarded as ``true parameters". Earlier works demonstrate its effectiveness on Maximum Entropy~\cite{chelba2006adaptation} or Support Vector Machine models~\cite{yang2007adapting,li2007regularized,aytar2011tabula}.

Despite that the notion of ``true parameters" of deep neural networks is intractable, several studies~\cite{bengio2012deep,yosinski2014transferable,oquab2014learning} have demonstrated the great transferability of representations trained on large scale datasets for general purpose. ~\cite{liu2019towards} theoretically studies properties of the pre-traind model and explains why it outperforms training from scratch. 
%These investigations give evidences to the popular solution of simply fine-tuning a pre-trained model with \LTWO\ regularization.

%These investigations give evidences to the most popular solution of simply fine-tuning a large scale dataset pre-trained model with the \LTWO\ regularization. Moreover, ~\cite{russakovsky2015imagenet} empirically studies which factors are essential for fine-tuning. ~\cite{liu2019towards} theoretically investigates some properties of deep transfer learning and explains why it outperforms training from scratch.
%This paper mainly discusses about inductive transfer learning, in which the source and target task are different tasks, and the target model is trained with labeled examples. Without disambiguation in this paper, we will use the name transfer learning for simpleness. Relevant research topics are domain adaptation and multi-task learning.

\subsection{Deep Inductive Transfer Learning}
%Fine-tuning a large scale dataset pre-trained model is the most simple and popular practice of transfer learning. Several earlier works~\cite{bengio2012deep,yosinski2014transferable,oquab2014learning} verify the transferability of features learned by large scale datasets for general purpose.~\cite{russakovsky2015imagenet} tries to discuss why ImageNet pre-trained model is so good for various downstreaming tasks. Recent work~\cite{liu2019towards} theoretically investigates some properties of deep transfer learning and explains why it outperforms training from scratch.

Despite the great transferability, recent work~\cite{li2018explicit} points out that naive fine-tuning the pre-trained model with \LTWO\ regularization may cause losses of the original knowledge. In order to overcome over-fitting, various transfer learning regularizers have been proposed. According to the type of regularized objectives, they can be categorized as parameter based~\cite{li2018explicit}, feature based~\cite{zagoruyko2016paying,yim2017gift,li2019delta} or spectral based~\cite{chen2019catastrophic} methods. 
%Recent work~\cite{li2018explicit} points out that naive fine-tuning the pre-trained model with \LTWO\ regularization may cause losses of the original knowledge. ~\cite{li2018explicit} solves this problem by extending the idea of shrinking towards true parameters into the deep learning paradigm and proposes to use the starting point as a reference for the feature extracting part of the whole model. They show that fine-tuning with \LTWOSP\ outperforms \LTWO\ normalization in evaluated tasks. ~\cite{aygun2017exploiting} suggests similar approach by regularizing the distribution of filters in convolutional layers. ~\cite{li2019delta} proposes \DELTA\, which instead uses the distance of deep representations between the source and target model as the regularization term. The regularization is improved by ``unactivated channel re-usage" to relax unnecessary constraining on features which are not discriminative for the target task. 

~\cite{chen2019catastrophic} improves the regularization of transfer learning from another angle. They propose Batch Spectral Shrinkage (BSS) to regularize spectral components of deep representations by penalizing small singular values. BSS is complementary to other regularizers. However, it doesn't deal with the issue of knowledge preserving.

%Besides regularizers, there are studies from other angles to improve inductive transfer learning, such as sample selection~\cite{ge2017borrowing,ngiam2018domain,jeon2020sample,li2020xmixup}, dynamic computing~\cite{guo2019spottune} and sparse transfer~\cite{wang2019pay}. 

Our paper adopts the general idea of preserving knowledge by regularizing features of the source model. While unlike previous methods, we do not directly use the original knowledge provided by the source model. Instead, we disentangle the useful part for reference to avoid negative transfer. There main differences are summarized in Table~\ref{table:compare}.

Studies from other angles, such as sample selection~\cite{ge2017borrowing,ngiam2018domain,jeon2020sample}, dynamic computing~\cite{guo2019spottune}, sparse transfer~\cite{wang2019pay} and cross-modality transfer~\cite{hu2020cross} are also important topics but out of this paper's scope. 

%Our improvement is based on~\cite{li2019delta} as we also perform representation alignments between the source and target model during fine-tuning. While the main difference is that, we create better representations as the reference. %Although the attention mechanism in~\cite{li2019delta} can be regarded as modifying the original representation by re-weighting each channel, this paper proposes a more powerful representation transformation by disentangling the positive part.
%we replace simple independent re-weighting of original channels by a careful representation disentanglement of the whole feature map, generating new representation which is more meaningful for the target task. 

\subsection{Representation Disentanglement}
The key assumption of representation disentanglement is that, a satisfactory representation should separate underlying factors of variations, which are 
compact, explanatory and independent with each other~\cite{goodfellow2009measuring,bengio2013representation}. Representation disentanglement has been widely applied in advanced generative algorithms such as Generative Adversarial Networks~\cite{goodfellow2014generative} and Variational Autoencoder~\cite{kingma2013auto}. Some works investigate general disentangling methods for data generating, including InfoGAN~\cite{chen2016infogan}, AC-GAN~\cite{odena2017conditional}, $\beta$-VAE~\cite{higgins2017beta}, FactorVAE~\cite{kim2018disentangling} and so on. %There are also some studies focus on specific tasks, such as image-to-image translation~\cite{gonzalez2018image} and age-invariant face recognition~\cite{wang2019decorrelated}.

Recently representation disentanglement is also demonstrated to be useful in tasks of unsupervised image-to-image translation and domain adaptation, which are more related to our work.~\cite{liu2018detach} proposes to 
%tackle the problem for cross-domain data and 
perform joint representation disentanglement and domain adaptation with only attribute supervision available in the source domain.~\cite{liu2018unified} further generalizes the previous study to a unified feature disentanglement network. %learning disentangled representation for multi-domain data. 
In this work, the data domain is first regarded as an interested underlying factor to be disentangled.
~\cite{peng2019domain} improves above studies by employing class disentanglement and minimizing the mutual information between disentangled features to enhance the disentanglement further.
%to boost the effect of adaptation, and enhancing the effect of disentanglement by minimizing the mutual information between them.

Our work is highly inspired and encouraged by the progress of domain information disentanglement~\cite{liu2018unified} and disentangling techniques~\cite{belghazi2018mine,peng2019domain}, while we are the first to utilize disentanglement methods to improve inductive transfer learning, aiming at (target) task specific feature disentanglement rather than domain invariant feature extraction in unsupervised domain adaptation. 

\section{Preliminaries}
\subsection{Problem Definition}
%In modern deep learning framework, the most important part is the feature extractor in charge of learning effective deep representations.
In inductive transfer learning, we are given a model pre-trained on the source task, with the parameter vector $\bm{\omega^0}$. For the desired task, the training set contains n tuples, each of which is denoted as 
$(\bm{x}_i,y_i)$. $\bm{x}_i$ and $y_i$ refers to the $i$-th example and its corresponding label. 

Let's further denote $z$ as the function of the neural network and $\bm{\omega}$ as the parameter vector of the target network. We have the objective of structural risk minimization
\begin{equation} \label{eq:srm}
  \min_{\bm{\omega}}\ \sum_{i=1}^n L(z( \bm{x}_i, \bm{\omega}), y_{i}) + \lambda\cdot\Omega(\bm{\omega},\bm{\omega}^0),
\end{equation}
where the first term is the empirical loss and the second is the regularization term. $\lambda$ is the coefficient to balance the effect between data fitting and reducing over-fitting.

\subsection{Regularizers for Transfer Learning}
%Regularization is critical for avoiding over-fitting, especially when we are offered a limited number of training examples. 
%Popular regularization choices such as $L^2$ or $L^1$ normalization are used to explicitly make the parameter small or/and sparse. While 
Recent studies in the deep learning paradigm show that SGD itself has the effect of implicit regularization that helps generalizing in over-parameterized regime~\cite{soltanolkotabi2018theoretical}. In addition, since fine-tuning is usually performed with a smaller learning rate and fewer epochs, it can be regarded as a form of implicit regularization towards the initial solution with good generalization properties~\cite{liu2019towards}. Besides, we give a brief introduction of state-of-the-art explicit regularizers for deep transfer learning.
%In transfer learning settings, since a pre-trained source model is provided, the target model is usually fine-tuned with a smaller learning rate and fewer number of epochs compared with training from scratch. This can be regarded as another form of implicit regularization towards the initial solution with good generalization properties~\cite{liu2019towards}. %So it is worth mentioning that fine-tuning without any explicit regularization terms is not a trivial baseline due to its implicit mechanism of meaningful regularizations. Beside this, we give a brief introduction of widely used and state-of-the-art explicit regularizers for inductive transfer learning. 

$\mathbf{L^2\ Penalty}$. The most common choice is the \LTWO\ penalty with the form of $\|\bm{\omega}\|^2_2$, also named weight decay in deep learning. %It tends to constrain the parameter around zero. 
From a Bayesian perspective, it refers to a Gaussian prior of the parameter with a zero mean. The shortcoming is that the meaningful initial point $\bm{\omega^0}$ is ignored. %~\cite{li2018explicit} argues that ${L^2}$ penalty is not adequate in transfer learning because the meaningful initial point $\bm{\omega^0}$ is ignored in the regularizer. \\

$\mathbf{L^2}$-$\mathbf{SP}$.~\cite{li2018explicit} follows the idea of shrinking towards chosen targets instead of zero. %In deep transfer learning scenarios, it differs that effective transfer of representations is considered as the objective rather than the classification module upon a fixed representation. 
They propose to use the starting point as the reference \\
\begin{equation}
\Omega(\bm{\omega}) = \frac{\alpha}{2}\|\bm{\omega_s}-\bm{\omega^0_s}\|^2_2 + 
\frac{\beta}{2}\|\bm{\omega_{\overline{s}}}\|^2_2, 
\end{equation}
where the first term refers to constraining the parameter of the part responsible for representation learning around the starting point, and the second is weight decay of the remaining part which is task specific. Since $\bm{\omega_{\overline{s}}}$ is general in all mentioned methods, we ignore it in following formulas. 

$\mathbf{DELTA}$.~\cite{li2019delta} extends the framework of feature distillation~\cite{Romero2014FitNetsHF,zagoruyko2016paying} by incorporating an attention mechanism. They constrain 2-d activation maps with respect to different channels by different strengths according to their values to the target task. Given a tuple of training example $(\bm{x}_i, y_i)$ and the distance metric between activation maps $\bm{D}$, the regularization is formulated as
%On top of regularizing feature maps of the source model, they constrain different channels by different strengths, according to their discriminative capacities for the target task. For a tuple of training example $(\bm{x}_i, y_i)$, we denote the 2-d feature map corresponding to the $j$-th channel of the target model as $FM_j(\bm{\omega_s}, \bm{x}_i$), and that of the source model as $FM_j(\bm{\omega_s^0}, \bm{x}_i$). Thus the distance between a feature map and its corresponding starting point is measured as
\iffalse
\begin{equation}
\bm{D}{^i_j} = 
\|
FM_j(\bm{\omega_s}, \bm{x}_i)-
FM_j(\bm{\omega_s^0}, \bm{x}_i)
%FM_j(\bm{\omega_s}, \bm{x}_i)-
%FM_j(\bm{\omega_s^0}, \bm{x}_i)
\|_2^2
\end{equation}
Then the regularization term of the representation learning part on all $C$ channels with respect to the $i$-th example is
\fi
\begin{equation} \label{eq:delta}
\begin{aligned}
\Omega{^i}(\bm{\omega_s}) =
 \frac{\alpha}{2}\sum_{j=1}^C 
 \mathrm{W}_{j}(\bm{\omega^0_s}, \bm{x}_i, y_i)\cdot\bm{D}{^i_j},
 %\\ &+ \beta \cdot\mathrm{Dist}(w,w^*).
  \end{aligned}
\end{equation}
where $C$ is the number of channels and $\mathrm{W}_{j}(\cdot)$ refers to the regularization weight assigned to the $j$-th channel. Specifically, each weight is independently evaluated by the performance drop when disabling that channel. 

$\mathbf{BSS}$. Authors in~\cite{chen2019catastrophic} propose Batch Spectral Shrinkage (\BSS), towards penalizing untransferable spectral components of deep representations. They figure out that spectral components which are less transferable are those corresponding to relatively small singular values. They apply differentiable SVD to compute all singular values of a feature matrix and penalize the smallest $k$ ones:
\begin{equation} \label{eq:bss}
\begin{aligned}
\Omega(\bm{\omega_s}) =
 \alpha \sum_{i=1}^k \sigma^2_{b+1-i},
  \end{aligned}
\end{equation}
where all singular values [$\sigma_1,\sigma_2...,\sigma_b$] are in the descending order. $\bm{\omega^0}$ is not involved as \BSS\ doesn't consider preserving the knowledge in the source model.
%the singular value decomposition of a feature matrix only relies on the current parameter of the target model. As stated by~\cite{chen2019catastrophic}, BSS mainly considers how to prevent from negative transfer instead of catastrophic forgetting. Therefore, BSS is orthogonal to existing fine-tuning methods. 

\begin{algorithm}[h]
\SetAlgoLined
\KwIn{labeled target dataset $\{(\bm{x}_i,y_i)\}_{i=1}^n$, model pre-trained on the source task $F_s$, representation disentangler $D$, classifier for disentangled representations $C$ and target model $F_t$.}
\KwOut{well-trained target model $\widehat{F}_t$.}

\tcp{\textbf{Representation Disentanglement}}
Set ($D$, $C$) trainable and $F_s$ frozen\;
\While{not converged}{
  Sample mini-batch $\mathcal{B}$: $\{(\bm{x}_j,y_j)\}_{j=1}^{\|\mathcal{B}\|}$\; 
  Calculate ($FM_{pos},FM_{neg}$) by Eq.~\ref{eq:di_forward} given ($\mathcal{B}, F_s, D$)\;
  %Set $M$ trainable and ($F_s$, $D$, $C$) frozen\;
  %Estimate mutual information between $f^c_{pos}$ and $f^c_{neg}$ by Eq.~\ref{eq:mi_optimization}\;
  %Calculate $L^D_{di}$ by Eq.~\ref{eq:di_mmd}\;
  %Calculate $L^D_{re}$ by Eq.~\ref{eq:di_re}\;
  %Calculate $L^D_{ce}$ by Eq.~\ref{eq:di_ce}\;
  Calculate $L^D_{di}$ by Max-MMD(Eq.~\ref{eq:di_mmd}) or Min-MI(Eq.~\ref{eq:di_mi})\;
  Calculate $L=L^D_{di}+L^D_{re}+L^D_{ce}$ by Eq.~\ref{eq:di_re}--\ref{eq:di_ce}\;
  Update $D$ and $C$ with SGD by minimizing $L$\;
 }
Obtain well-trained $\widehat{D}$ = $D$\;

\tcp{\textbf{Fine-tuning}}
Set ($F_s$, $\widehat{D}$) frozen and $F_t$ trainable\;
\While{not converged}{
  Sample mini-batch $\mathcal{B}$: $\{(\bm{x}_j,y_j)\}_{j=1}^{\|\mathcal{B}\|}$\; 
  Obtain $FM_{pos}$ by Eq.~\ref{eq:di_forward} given ($\mathcal{B}, F_s, \widehat{D}$)\;
  Calculate $\Omega(\bm{\omega_s})=\sum_{j=1}^{\|\mathcal{B}\|}\Omega{^j}(\bm{\omega_s})$ by Eq~\ref{eq:pddr_reg}\;
  Update $F_t$ by Eq~\ref{eq:srm}\;
}
\KwRet{$\widehat{F_t} = F_t$}
\caption{The framework of \TheName}
\label{implement}
\end{algorithm}

\section{Target-awareness Disentanglement}
Features extracted from the source model, which is usually pre-trained over a large scale dataset with diverse categories, are often noisy for a specific target task. Irrelevant ingredients w.r.t. the target task contained the general knowledge may lead to negative transfer. Our aim is to disentangle the positive ingredients (relevant to the target task) from the entire representation. To achieve this goal, three conditions: distinguishable, discriminative, and recoverable should be satisfied:
\begin{itemize}
\item The positive part should be \emph{distinguishable} from the negative part. In other words, two features with similar patterns or semantic relations should not be disentangled into different parts. This is the most crucial component in this framework. 
\item The positive part should be \emph{discriminative} on the target task. The aforementioned disentangler is usually symmetric, i.e. we can define either part as the ``positive'' if without external supervision. Therefore, an extra signal is needed to discriminate the positive part. 
\item The original representation should be \emph{recoverable} by the disentangled parts. For a disentangling operation, both the disentangled parts should not represent knowledge beyond the original representation. Otherwise, the transformation may result in non-generalizable features. 
\end{itemize}

\subsection{General definitions and notations}.
We first specify general definitions and notions used in our algorithm. Different with the main stream of disentanglement studies which try to separate various atomic attributes such as the color or angle, we care about disentangling components relevant to the target task from the whole representation produced by the source model.
%, the only interested output in transfer learning scenarios is the target task. Therefore, we 
%We use a simple module composed of a 1x1 convolutional, batch normalization and ReLU layer. to 
Formally, we disentangle the original representation $FM_{ori} \in R^{C \times H \times W}$ obtained from the pre-trained model into the positive and negative part with the disentangler module $D$:
\begin{equation} \label{eq:di_forward}
FM_{pos}, FM_{neg} = D(FM_{ori}),
\end{equation}
where $FM_{pos}$ and $FM_{neg}$ have the same shape with $FM_{ori}$. For efficient estimation and optimization of the disentanglement, we further denote the mapping functions $\mathcal{F^C} : R^{C \times H \times W} \to R^{C}$ and $\mathcal{F^S} : R^{C \times H \times W} \to R^{H \times W}$, representing dimension reduction along the spatial and channel direction respectively. Therefore we get
\begin{equation}
f^c_* = \mathcal{F^C}(FM_*), f^s_* = \mathcal{F^S}(FM_*),
%\mathrm{f^a}_* = \mathcal{F^A}(FM_*)
\end{equation}
where $*$ refers to either $pos$ or $neg$.

\subsection{Feature Disentanglement}
The design of our disentangling is motivated by the following two assumptions, to which the ideally disentangled relevant (positive) and irrelevant (negative) part are expected to conform. 

\begin{itemize}
\item \emph{Non-overlapped Visual Attention.} The visual attention can be regarded as a most interpretable kind of knowledge learned by DNNs~\cite{zagoruyko2016paying}. Intuitively, the disentangled two parts should pay attention to different visual regions. Otherwise, similar patterns are more likely to concurrently exist in different parts, implying an incomplete disentanglement. 
\item \emph{Independent Semantic Representation.} Another common assumption is that, patterns relevant to a same target object are likely to have dependence with each other, but not vice versa. For example, in order to recognize different dogs, a pattern on eyes often implies a pattern on noses, but not much likely to imply a pattern on indoor objects such as cups or floors. 
\end{itemize}
Both aforementioned assumptions can be demonstrated by the example in Figure~\ref{fig:arch}. Next we describe the two methods in details. 

%The positive part is distinguished by a simple discriminator. 
%The disentangled positive part is expected to filter useless knowledge with respect to the target task, which may hurt fine-tuning if used as the regularizer. 
%Then we fine-tune the target model with the regularization to restrict the distance between feature maps and the corresponding disentangled ones. A framework of the approach is illustrated in Figure~\ref{fig:arch}. We explain the components in the following paragraphs.

%Our work follows the general idea of using the starting point as the reference to overcome catastrophic forgetting. Aiming at the problem of negative transfer, we propose a straightforward way that performs an explicit representation transformation adapted for the target task. Specifically, it is a two-step algorithm. First we disentangle the positive part from the original representation provided by the pre-trained model. Then we fine-tune the target model with the regularization to restrict the distance between feature maps and the corresponding disentangled feature maps. A framework of the approach is illustrated as Figure~\ref{fig:arch}. We explain the components in the following paragraphs.

\textbf{Non-overlapped Visual Attention with Max-MMD}.
Imitating the visual attention mechanism of humans, we force the two parts to pay attention to different spatial regions within the original image. This is achieved by enlarging their statistical distributions along the spatial dimension. We use the Maximum Mean Discrepancy (MMD) to measure the distribution distance between the two spatial representations. Maximum Mean Discrepancy (MMD) is originally designed to test whether two distributions are the same~\cite{gretton2012kernel}. Further, it's also widely used as a criterion to measure the distance of two distributions in domain adaptation tasks~\cite{pan2010domain,tzeng2014deep,long2017deep} and 
generative adversarial networks~\cite{sutherland2016generative,arbel2018gradient}. Under a commonly adopted Reproducing Kernel Hilbert Space (RKHS) assumption, the MMD can be represented as an unbiased approximation with the kernel form as follows. 

Denoting $X_s=\{x_s^1, x_s^2,...,x_s^n\}$ and $X_t=\{x_t^1, x_t^2,...,x_t^m\}$ as random variable sets with distributions $P$ and $Q$, an empirical estimate~\cite{tzeng2014deep,long2015learning} of the MMD between $P$ and $Q$ compares the square distance between the empirical kernel mean embeddings as
\begin{equation} 
\begin{aligned}
\mathbf{MMD}(P, Q) & = \| \frac{1}{m}\sum_{i=1}^m k(x_s^i) - \frac{1}{n}\sum_{j=1}^n k(x_t^j) \|^2,
  \end{aligned}
\end{equation}
where $k$ refers to the kernel, as which a Gaussian radial basis function (RBF) is usually used in practice~\cite{long2015learning,louizos2016variational}. 

\iffalse
\begin{equation} 
\begin{aligned}
\mathbf{MMD}(P, Q) & = \frac{1}{n^2}\sum_{i=1}^n \sum_{j=1}^n k(x_s^i, x_s^j) \\
& + \frac{1}{m^2}\sum_{i=1}^m \sum_{j=1}^m k(x_t^i, x_t^j) \\
& - \frac{2}{nm}\sum_{i=1}^n \sum_{j=1}^m k(x_s^i, x_t^j) 
  \end{aligned}
\end{equation}
\fi

Our objective is to enlarge the MMD between the disentangled positive and negative part along the spatial dimension. Intuitively, this would explicitly encourage these two parts to recognize different regions of the input image. For stabler optimization, we minimize the negative exponent of the MMD as followed:
\begin{equation} \label{eq:di_mmd}
\begin{aligned}
L^D_{di}=\lambda_{di}e^{-\mathbf{MMD}(f^s_{pos}, f^s_{neg})}.
  \end{aligned}
\end{equation}

\textbf{Independent Semantic Representation with Min-MI}.
In information theory, mutual information (MI) between two random variables quantifies the amount of information obtained about one through observing the other. 
Formally, the mutual information between random variables $X$ and $Z$ is defined as
\begin{equation} \label{eq:mi_def}
\begin{aligned}
I(X;Z) = D_{KL}(\mathbb{P}_{XZ}\|\mathbb{P}_X \otimes \mathbb{P}_Z)
  \end{aligned}
\end{equation}
, where $\mathbb{P}_{X}$ and $\mathbb{P}_{Z}$ are the marginal distributions, and $\mathbb{P}_{XZ}$ is their joint distribution. In general, $I(X;Z)$ is non-negative and zero only when $X$ and $Z$ are independent.

Recent study~\cite{higgins2017beta} demonstrated that disentanglement of interested factors can be enhanced by encouraging independence between them. Further, \cite{peng2019domain} practically minimized the mutual information between disentangled features to strengthen class/domain disentanglement. Inspired by ~\cite{peng2019domain}, to achieve the goal of disentangling the original representation into two parts which are both meaningful and complementary, we minimize the mutual information between probability distributions of $f^c_{pos}$ and $f^c_{neg}$.

Since the exact computation of mutual information for high-dimensional data is rather hard, we leverage recent approach for efficient estimation of mutual information~\cite{belghazi2018mine} by a neural network $T_\theta$:
\begin{equation} \label{eq:mi_optimization}
\begin{aligned}
I(\widehat{X;Z})_n = \sup_{\theta \in \Theta}\mathbb{E}_{\mathbb{P}_{XZ}^{(n)}}[T_\theta]-\mathrm{log}(\mathbb{E}_{\mathbb{P}_X^{(n)} \otimes \mathbb{\hat{P}}_Z^{(n)}}[e^{T_\theta}]).
  \end{aligned}
\end{equation}

Denoting the optimal solution as $\hat{\theta}$, the mutual information can be computed by Monte-Carlo approximation for a mini-batch of $b$ examples as
\begin{equation} \label{eq:mi_estimation}
\begin{aligned}
I(X;Z;\hat{\theta})_b = \frac{1}{b}\sum_{i=1}^b T(x^i,z^i,\hat{\theta}) - \mathrm{log}(\frac{1}{b}\sum_{i=1}^b e^{T(x^i,\overline{z}^i,\hat{\theta})})
\end{aligned}
\end{equation}
, where each $(x^i, z^i)$ is drawn from the joint distribution $\mathbb{P}_{XZ}$, and $\overline{z}^i$ is drawn from the marginal distribution $\mathbb{P}_Z$.

Mutual information estimation by Eq~\ref{eq:mi_optimization} is a maximizing problem by optimizing on $T_{\theta}$, while our objective is to minimize the mutual information by learning the disentangled representations. Therefore, we adopt common practice to alternatively update $T_{\theta}$ and $D$. Given $T_{\hat{\theta}}$ estimated for the current distributions of $f^c_{pos}$ and $f^c_{neg}$, we update $D$ with
\begin{equation} \label{eq:di_mi}
\begin{aligned}
L^D_{di}=\lambda_{di}
I(f^c_{pos};f^c_{neg};\hat{\theta})_b
  \end{aligned}
\end{equation}

\begin{table*}[t]
\centering
\caption{Comparison of top-1 accuracy (\%) with different methods. Baselines are No Regularization, \LTWO, \BSS~\cite{chen2019catastrophic},  \LTWOSP~\cite{li2018explicit},  \AT~\cite{zagoruyko2016paying} and \DELTA~\cite{li2019delta}. \TheName-MMD refers to \TheName\ with Max-MMD and \TheName-MI refers to \TheName\ with Min-MI.}
\label{table:performance}
\begin{tabular}{l|cccccccc}
Dataset & No Reg. & \LTWO\ & \BSS\ & \LTWOSP\ & \AT\ & \DELTA\ & \TheName-MMD & \TheName-MI \\\hline
CUB-200-2011 & 79.31$\pm$0.20 & 78.96$\pm$0.14 & 79.52$\pm$0.13 & 79.51$\pm$0.23 & 80.34$\pm$0.09 & 80.93$\pm$0.11 & 82.07$\pm$0.07 & \textbf{82.93$\pm$0.18} \\
Stanford-Dogs & 84.47$\pm$0.08 & 84.56$\pm$0.08 & 84.74$\pm$0.13 & 90.16$\pm$0.09 & 87.88$\pm$0.05 & 89.83$\pm$0.05 & 90.36$\pm$0.14 & \textbf{90.80$\pm$0.07}
\\
Stanford-Cars & 88.62$\pm$0.12 & 88.65$\pm$0.23 & 89.78$\pm$0.02 & 89.40$\pm$0.03 & 88.03$\pm$0.08 & 89.12$\pm$0.01 & 90.02$\pm$0.07 & \textbf{90.64$\pm$0.12} \\
Flower-102 & 89.43$\pm$0.08 & 89.72$\pm$0.29 & 89.71$\pm$0.1 & 89.87$\pm$0.33 & 89.08$\pm$0.08 & 89.76$\pm$0.17 & 91.34$\pm$0.23 & \textbf{91.74$\pm$0.12} \\
MIT-Indoor & 82.20$\pm$0.15 & 81.94$\pm$0.41 & 82.31$\pm$0.3 & 84.89$\pm$0.11 & 85.23$\pm$0.07 & 85.53$\pm$0.34 & \textbf{85.79$\pm$0.15} & 85.56$\pm$0.23\\
Texture & 68.32$\pm$0.24 & 68.22$\pm$0.22 & 68.65$\pm$0.38 & 69.68$\pm$0.59 & 69.78$\pm$0.22 & 70.85$\pm$0.27 & \textbf{71.58$\pm$0.52} & 71.51$\pm$0.44 \\
Oxford-Pet & 93.75$\pm$0.13 & 93.7$\pm$0.25 & 93.91$\pm$0.15 & 93.94$\pm$0.2 & 94.09$\pm$0.05 & 93.90$\pm$0.24 & 94.37$\pm$0.15 & \textbf{94.43$\pm$0.10} \\
%Food101-30 & 63.48$\pm$0.01 & 63.30$\pm$0.22 & 63.78$\pm$0.15 & 63.20$\pm$0.33 & 63.88$\pm$0.09 & 64.37$\pm$0.16 & \textbf{65.53$\pm$0.14} \\
%Food101-150 & 77.45$\pm$0.02 & 77.68$\pm$0.23 & 77.59$\pm$0.20 & 77.78$\pm$0.08 & 78.40$\pm$0.08 & 78.21$\pm$0.15 & \textbf{79.14$\pm$0.11} \\
\hline
%Average & 80.78 & 80.75 & 81.11 & 82.05 & 81.86 & 82.50 & 83.36 \hline
Average & 83.73 & 83.67 & 84.09 & 85.35 & 84.91 & 85.70 & 86.51 & \textbf{86.80} \\
\hline
\end{tabular}
%\vspace{-3mm}
\end{table*}

\subsection{Reconstruction Requirement}.
As both the positive and negative part are trained by the flexible disentangler, it is easy to produce two parts of meaningless representations with the only objective of disentangling with Max-MMD or Min-MI. To ensure the disentanglement is restricted within the knowledge of the source model rather than an arbitrary transformation, we add the reconstruction requirement to constrain the disentangled results. Specifically, the disentangled positive and negative parts are required to be capable of recovering the original representation by point-wise addition:
%additional restriction is required to ensure that the disentanglment makes sence rather than arbitrary transformation.
%A major concern about this algorithm may be that how to guarantee the disentanglement is informative rather than arbitrary transformation, since both the positive and negative part are trained by the flexible disentangler. We argue that the effect is protected by two mechanisms. One is that we use a lightweight module for disentangling, which is consisted of a convolutional layer followed by a batch normalization and nonlinear activation layer. The other is that the disentangled parts are required to be able to reconstruct the original representation by directly point-wise addition:
\begin{equation} \label{eq:di_re}
\begin{aligned}
L^D_{re}=\lambda_{re}\|FM_{pos} + FM_{neg} - FM_{ori} \|^2_2 .
  \end{aligned}
\end{equation}

\subsection{Distinguishing the Positive Part}.
Since above representation disentanglement is actually symmetry for each part, an explicit signal is required to distinguish features which are useful for the target task. In particular, the selected layer for representation transfer is followed by a classifier consisting of a global pooling layer and a fully connected layer sequentially. A regular cross entropy loss is added to explicitly drive the disentangler to extract into the positive part components which are discriminative for the target task. Denoting the involved classifier as $C$, we have
\begin{equation} \label{eq:di_ce}
\begin{aligned}
L^D_{ce}=\lambda_{ce}\mathbf{CrossEntropy}(C(f^c_{pos}), y_i) .
  \end{aligned}
\end{equation}

\subsection{Regularizing the Disentangled Representation}.
After the step of representation disentanglement, we perform fine-tuning over the target task. We regularize the distance between a feature map and its corresponding starting point. Quite different from previous feature map based regularizers as~\cite{Romero2014FitNetsHF,zagoruyko2016paying,li2019delta}, the starting point here is the disentangled positive part of the original representation. 
%Another difference is that we regularize each channel equally since the whole representation has been adapted for the target task. 
The regularization term corresponding to some example ($\bm{x}_i, y_i$) becomes:
\begin{equation} \label{eq:pddr_reg}
\Omega{^i}(\bm{\omega_s}) = 
\frac{\alpha}{2}\|
FM(\bm{\omega_s}, \bm{x}_i)-
FM_{pos}(\bm{\omega_s^0}, \bm{\omega_{di}}, \bm{x}_i)
\|_2^2,
\end{equation}
where $\bm{\omega_{di}}$ refers to the parameter of the disentangler $D$ which is frozen during the fine-tuning stage. The complete training procedure is presented in Algorithm~\ref{implement}.

\iffalse
\begin{table}[ht]
\centering
\caption{Characteristics of the target datasets: name, the number of category and the average number of training examples per category.} 
\label{table:dataset}
\begin{tabular}{l|c|c}
Name & \# categories & \# training \\\hline
Stanford Dogs & 120 & 100 \\
Stanford Cars & 196 & $\sim$42 \\
CUB-200-2011 & 200 & 30 \\
Flower-102 & 102 & 10 \\
DTD & 47 & 40 \\
MIT Indoors & 67 & $\sim$83 \\
Food-101(subset) & 101 & 30 \\
FGVC-Aircraft & 102 & 66
\end{tabular}
\end{table}
\fi

\iffalse
\begin{table}[h]
\centering
\caption{Top-1 accuracy (\%) of \TheName\ and best baseline method in Table~\ref{table:performance} combined with \BSS~\cite{chen2019catastrophic}.}
\label{table:bss}
\begin{tabular}{l|cc}
Dataset & BestBase+\BSS\ & \TheName+\BSS \\\hline
MIT indoor-67 & 85.37$\pm$0.06 & 85.88$\pm$0.28 \\
Flower-102 & 90.32$\pm$0.12 & 90.82$\pm$0.22 \\
Stanford Cars & 88.05$\pm$0.20 & 88.61$\pm$0.18 \\
DTD & 71.31$\pm$0.30 & 71.68$\pm$0.31 \\
CUB-200-2011 & 79.30$\pm$0.13 & 81.18$\pm$0.26 \\ 
Food-101 & 61.44$\pm$0.13 & 64.15$\pm$0.17 \\
FGVC Aircraft & 82.50$\pm$0.35 & 83.46$\pm$0.44 \\
Stanford Dogs & 90.89$\pm$0.13 & 90.88$\pm$0.07 \\
\end{tabular}
\vspace{-2mm}
\end{table}
\fi

\section{Experiments}

\begin{table}[t]
\centering
\caption{Comparison of top-1 accuracy (\%) on CUB-200-2011 with respect to different sampling rates.}
\begin{threeparttable}
\begin{tabular}{l|ccc}
\multirow{2}*{Algorithm} & \multicolumn{3}{c}{Sampling Rate} \\
 & 50\% & 30\% & 15\% \\\hline
\LTWO & 70.44 & 60.64 & 34.98 \\
\BSS & 71.09 & 62.31 & 38.36 \\
\LTWOSP & 70.37 & 60.93 & 36.71  \\
\AT & 71.85 & 59.87 & 35.84 \\
\DELTA & 71.96 & 64.22 & 38.19 \\
\TheName-MMD & 73.84 & 65.47 & 41.77 \\
\TheName-MI & \textbf{75.46} & \textbf{67.42} & \textbf{42.07} \\
\end{tabular}
\end{threeparttable}
\label{table:perf_cub}
\vspace{-4mm}
\end{table}

\subsection{Datasets}
We select several popular transfer learning datasets to evaluate the effectiveness of our method.

\textbf{Stanford Dogs.} 
The Stanford Dogs~\cite{KhoslaYaoJayadevaprakashFeiFei_FGVC2011} dataset consists of images of 120 breeds of dogs, each of which containing 100 examples used for training and 72 for testing. It's a subset of ImageNet. %Bounding box annotations are not used in all experimens.

\textbf{MIT Indoor-67.} MIT Indoor-67~\cite{quattoni2009recognizing} is an indoor scene classification task consisting of 67 categories. There are 80 images for training and 20 for testing for each category. 

\textbf{CUB-200-2011.} Caltech-UCSD Birds-200-2011~\cite{WelinderEtal2010} contains 11,788 images of 200 bird species from around the world. Each species is associated with a Wikipedia article and organized by scientific classification. 

%\textbf{Food-101.} Food-101~\cite{bossard14} is a large scale data set consisted of more than 100k food images divided into 101 different kinds. To better fit transfer learning applications, we use two subsets which contains 30 and 150 training examples per category respectively.

\textbf{Flower-102.} Flower-102~\cite{Nilsback2008Automated} consists of 102 flower categories. 1020 images are used for training and 6149 images for testing. Only 10 samples are provided for each category during training.

\textbf{Stanford Cars.} The Stanford Cars~\cite{KrauseStarkDengFei-Fei_3DRR2013} dataset contains 16,185 images of 196 classes of cars. The data is split into 8,144 training and 8,041 testing images.

\textbf{Oxford-IIIT Pet.} The Oxford-IIIT Pet~\cite{parkhi2012cats} dataset is a 37-category pet dataset with about 200 cat or dog images for each class.

%\textbf{FGVC-Aircraft.} FGVC-Aircraft~\cite{Maji2013FineGrainedVC} is a fine-grained visual classification dataset composed of more than 10,000 images of aircraft across 102 different aircraft models.  

\textbf{Textures.} Describable Textures Dataset~\cite{cimpoi14describing} is a texture database, containing 5640 images organized by 47 categories according to perceptual properties of textures.

%We summarize all these datasets in~\ref{table:dataset}.

\subsection{Settings and Hyperparameters}
We implement transfer learning experiments based on ResNet~\cite{he2016deep}. For MIT indoor-67, we use ResNet-50 pre-trained with large scale scene recognition dataset Places 365~\cite{zhou2017places} as the source model. For remaining datasets, we use ImageNet pre-trained ResNet-101 as the source model. Input images are resized with the shorter edge being 256 and then random cropped to $224\times224$ during training.

For optimization, we first train 5 epochs to optimize the disentangler by Adam with the learning rate of 0.01. All involved hyperparameters are set to default values of $\lambda_{di}=10^{-2}, \lambda_{ce}=10^{-3}, \lambda_{re}=10^{-2}$. Then we use SGD with the momentum of 0.9, batch size of 64 and initial learning rate of 0.01 for fine-tuning the target model. We train 40 epochs for each dataset. The learning rate is divided by 10 after 25 epochs. 
%For the disentangler we adopt Adam with the learning rate of 0.01 as the optimizer. We train only 5 epochs to optimize the disentangler. 
We run each experiment three times and report the average top-1 accuracy. 

\TheName\ (both with Max-MMD and Min-MI) is compared with state-of-the-art transfer learning regularizers including \LTWOSP~\cite{li2018explicit}, \AT~\cite{zagoruyko2016paying}, \DELTA~\cite{li2019delta} and \BSS~\cite{chen2019catastrophic}. We perform 3-fold cross validation searching for the best hyperparameter $\alpha$ in each experiment. For \LTWOSP, \DELTA\ and \TheName, the search space is [$10^{-3}$, $10^{-2}$, $10^{-1}$]. Although authors in \AT\ and \BSS\ recommended fixed values of $\alpha$ ($10^3$ for \AT\ and $10^{-3}$ for \BSS), we also extend the search space to [$10^2$, $10^3$, $10^4$] for \AT\ and [$10^{-4}$, $10^{-3}$, $10^{-2}$] for \BSS. 

\begin{figure*}[ht]
\subfigure[Input]
{
	\begin{minipage}{1.9cm}
    \center
	\includegraphics[scale=0.45]{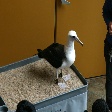}  
	\includegraphics[scale=0.45]{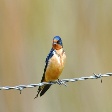}  
	\includegraphics[scale=0.45]{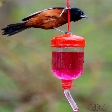} \end{minipage}
}\subfigure[Origin]
{
	\begin{minipage}{1.9cm}
    \center
	\includegraphics[scale=0.45]{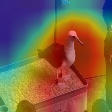}
	\includegraphics[scale=0.45]{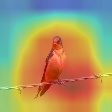}
	\includegraphics[scale=0.45]{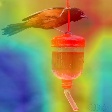}
	\end{minipage}
}\subfigure[Positive]
{
	\begin{minipage}{1.9cm}
    \center
	\includegraphics[scale=0.45]{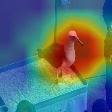}
	\includegraphics[scale=0.45]{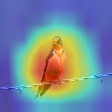}
	\includegraphics[scale=0.45]{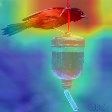}  
	\end{minipage}
}\subfigure[Negative]
{
	\begin{minipage}{1.9cm}
    \center
	\includegraphics[scale=0.45]{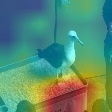}  \includegraphics[scale=0.45]{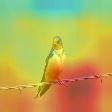}  \includegraphics[scale=0.45]{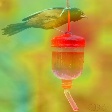}
	\end{minipage}
}
\setcounter{subfigure}{0}
\subfigure[Input]
{
	\begin{minipage}{1.9cm}
    \center
	\includegraphics[scale=0.45]{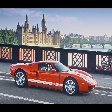}  \includegraphics[scale=0.45]{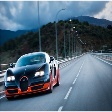}  \includegraphics[scale=0.45]{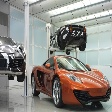} \end{minipage}
}\subfigure[Origin]
{
	\begin{minipage}{1.9cm}
    \center
	\includegraphics[scale=0.45]{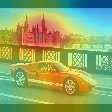} 	\includegraphics[scale=0.45]{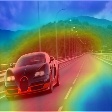} 	\includegraphics[scale=0.45]{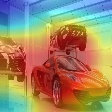}
	\end{minipage}
}\subfigure[Positive]
{
	\begin{minipage}{1.9cm}
    \center
	\includegraphics[scale=0.45]{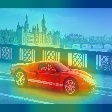}  	\includegraphics[scale=0.45]{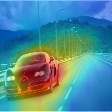}  	\includegraphics[scale=0.45]{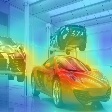}  
	\end{minipage}
}\subfigure[Negative]
{
	\begin{minipage}{1.9cm}
    \center
	\includegraphics[scale=0.45]{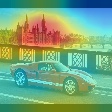}  	\includegraphics[scale=0.45]{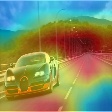}
	\includegraphics[scale=0.45]{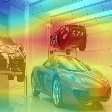}
	\end{minipage}
}
\vspace{-2mm}
\caption{The effectiveness of representation disentanglement on CUB-200-2011 (left) and Stanford Cars (right). For each dataset, we select three typical cases for demonstration. In addition to the input image (a), we add spatial attention map onto the original image in column (b), (c), and (d) using the input image and the desired representation of the last convolutional layer of ResNet-101. Specifically, (b) is the original representation generated by the ImageNet pre-trained model. (c) and (d) are the disentangled positive and negative part by \TheName.}
\label{fig:showcase}
\vspace{-3mm}
\end{figure*}

\begin{figure}[t]
%\centering
\subfigure[Flower-102 Original]{
%\begin{minipage}[t]{0.45\linewidth}
\includegraphics[width=1.5in]{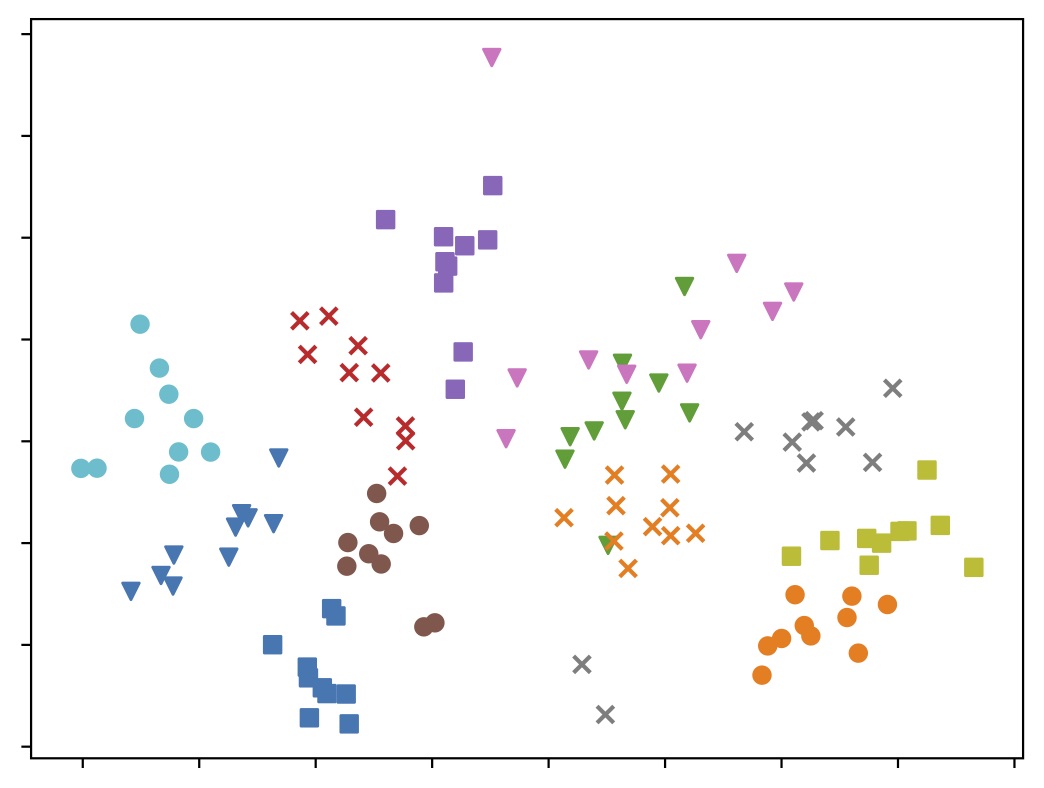}
%\end{minipage}%
}%
\subfigure[Flower-102 Disentangled]{
%\begin{minipage}[t]{0.45\linewidth}
\includegraphics[width=1.5in]{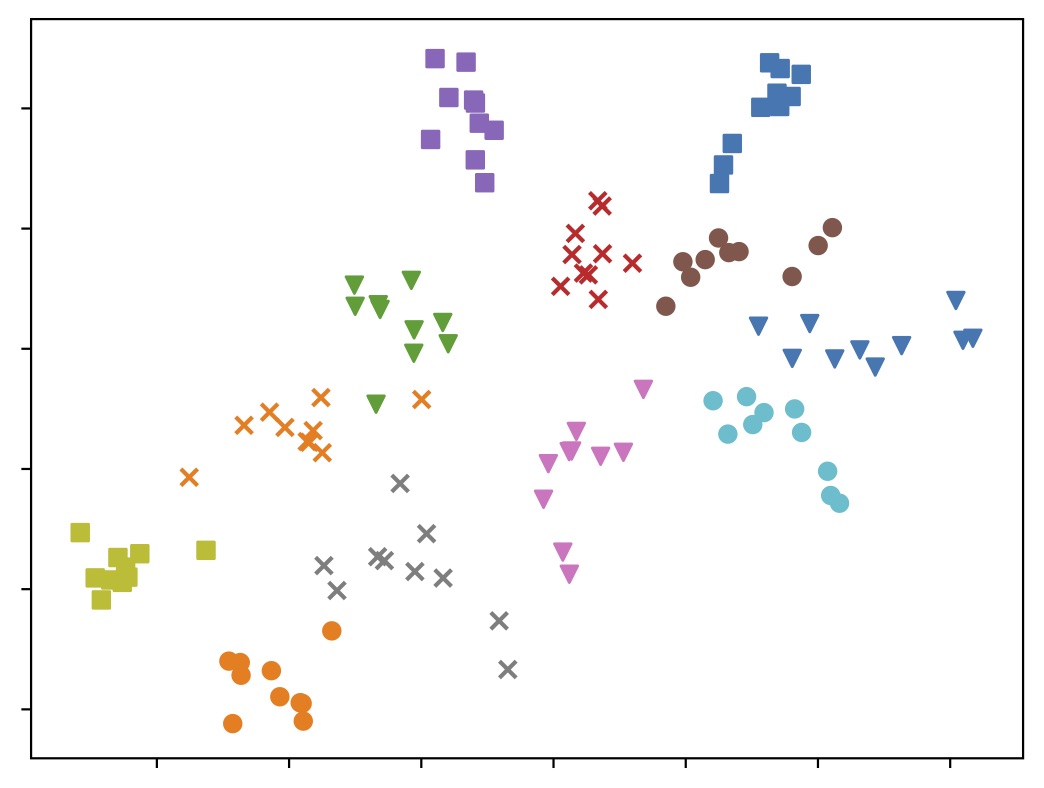}
%\end{minipage}%
}%
\quad
\subfigure[Indoor-67 Original]{
%\begin{minipage}[t]{0.45\linewidth}
\includegraphics[width=1.5in]{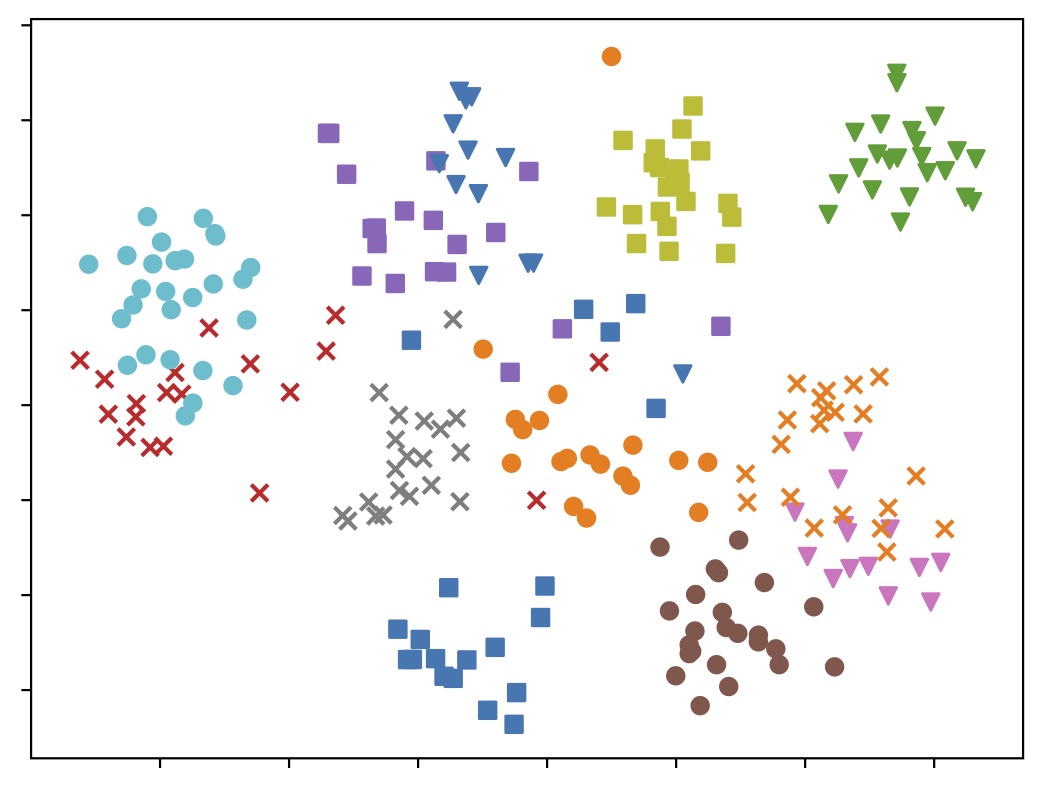}
%\end{minipage}
}%
\subfigure[Indoor-67 Disentangled]{
%\begin{minipage}[t]{0.45\linewidth}
\includegraphics[width=1.5in]{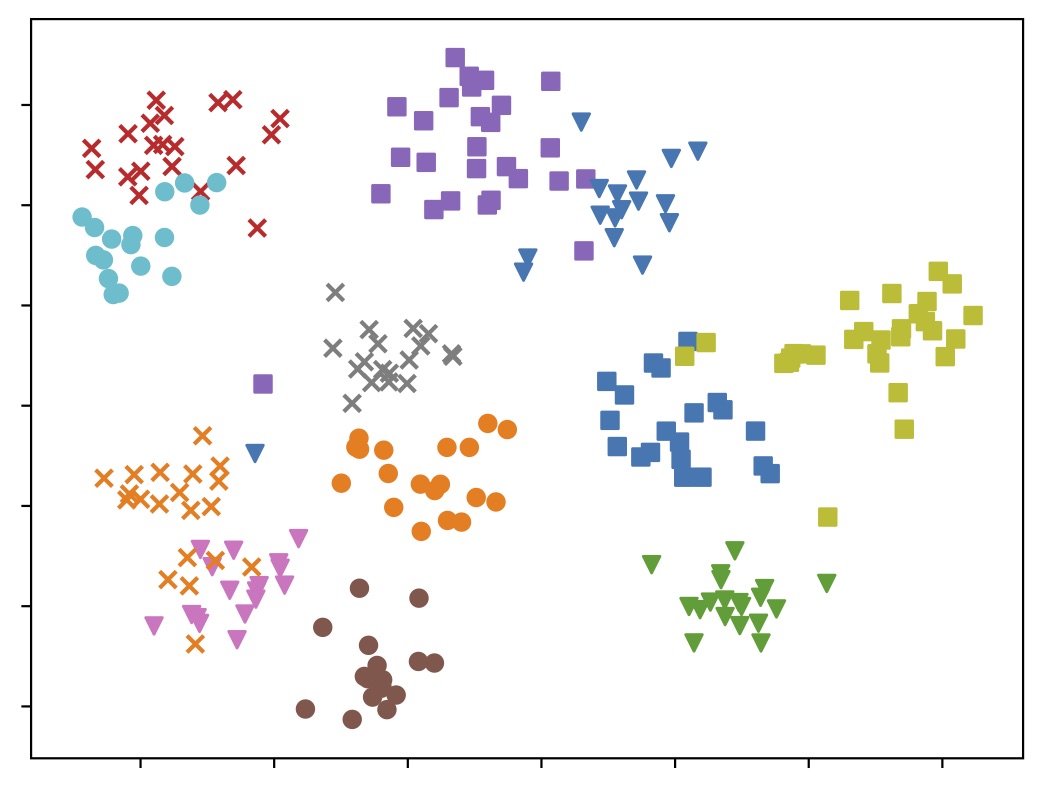}
%\end{minipage}
}%
\vspace{-2mm}
\caption{Visualization of the original (a, c) and disentangled (b, d) feature representations by t-SNE. Different colors and markers are used to denote different categories.}
\label{fig:tsne}
\vspace{-4mm}
\end{figure}

\subsection{Results}
%1.transfer learning hardly benefits from regularizations of non-informative piror
%2.informative piror brings significant difference which believes the original knowledge contained in the source model. regularization on representations performs more robust to negative transfer than parameters.
%3.\TheName\ consistently outperforms DELTA, especially  for datasets suffering from negative transfer
%4.orthogonal with bss.

While recent theoretical studies proved that weight decay actually has no regularization effect~\cite{van2017l2,golatkar2019time} when combined with common used batch normalization, we use \emph{No Regularization} as the most naive baseline, reevaluating \LTWO. From Table~\ref{table:performance} we observe that \LTWO\ does not outperform fine-tuning without any regularization. This may imply that deep transfer learning hardly benefits from regularizers of non-informative priors.% like \LTWO.

Advanced works~\cite{zagoruyko2016paying,li2018explicit,li2019delta} adopt regularizers using the starting point of the reference for knowledge preserving. From the perspective of Bayes theory, these are equivalent to the informative prior which believes the knowledge contained in the source model, in the form of either parameters or behavior. Table~\ref{table:performance} shows that these algorithms obtain significant improvements on some datasets such as Stanford Dogs and MIT indoor-67, where the target dataset is very similar to the source dataset. However, benefits are much less on other datasets such as CUB-200-2011, Flower-102 and Stanford Cars. %\DELTA\ usually obtains greater improvements on tasks benefiting from \LTWOSP. While regarding tasks where \LTWOSP\ performs worse than \LTWO\, e.g. Stanford Cars and FGVC Aircraft, the improvement of \DELTA\ is also negative or very marginal.

Table~\ref{table:performance} illustrates that \TheName\ consistently outperforms all above baselines over all evaluated datasets. It outperforms naive fine-tuning regularizer \LTWO\ by more than 2\% on average. Except for Stanford Dogs and MIT Indoor-67, improvements are still obvious even compared with state-of-the-art regularizers \LTWOSP, \AT, \DELTA\ and \BSS. %Besides, we do additional experiments on \textbf{portions} of the training dataset, showing that our algorithms also \textbf{significantly} outperforms competitors. The detailed results are presented in the supplementary material. %Specifically, \TheName\ improves by a large margin on tasks where \LTWOSP\ and \DELTA\ perform bad. For those target tasks on which \LTWOSP\ or \DELTA\ performs better than non-informative regularizers, \TheName\ still exhibits some advantage. %It is worth noting that the \TheName\ is even superior to the best of baseline algorithms combined with \BSS\ in Table~\ref{table:bss}. Although \TheName\ does not obviously benefit from combining \BSS, it is still superior to the best baseline choice of No Regularization, \LTWO\, \LTWOSP\ and \DELTA\ for most datasets. 

%We will present more analysis in Section \ref{sec5.3} to explain which kinds of datasets will benefit more from \TheName.

To evaluate the scalability of our algorithm with more limited data, we conduct additional experiments on subsets of the standard dataset CUB-200-2011. Baseline methods include \LTWO, \BSS~\cite{chen2019catastrophic},  \LTWOSP~\cite{li2018explicit},  \AT~\cite{zagoruyko2016paying} and \DELTA~\cite{li2019delta}. Specifically, we random sample 50\%, 30\% and 15\% training examples for each category to construct new training sets. Results show that our proposed \TheName\ achieves remarkable improvements compared with all competitors, as presented in Table~\ref{table:perf_cub}. We will provide more explanations about the results in the supplementary material. 

%Figure~\ref{fig:samplerate} shows how these methods behave when reducing the size of the training set. For clear illustration, we treat all regularizers only considering the risk the \emph{catastrophic forgetting} as a group, namely SPAR as they are all follow the framework of using the Starting Point As the Reference. The average accuracy of \LTWOSP, \AT\ and \DELTA\ is used to represent for SPAR. \BSS, which is designed for suppressing the untransferable ingredients of features, only tackles the problem of \emph{negative transfer}. While \TheName\ deals with \emph{both} challenges. We plot the improvements of these methods compared with naive fine-tuning with \LTWO\ regularization. We can observe from Figure~\ref{fig:samplerate} that \BSS\ and \TheName\ obtain increased improvements as the sampling rate reduces, implying that the negative impact from the source model is greater when the target dataset is smaller. Although sharing the same trend, \TheName\ always outperform \BSS\ with an obvious margin at all sampling rates. While the curve of SPAR is much stabler as the sampling rate reduces. 

\section{Discussions}
In this section, we dive deeper into the mechanism and experiment results to explain why target-awareness disentanglement provides better reference. In subsection \emph{Representation Visualization}, we show the effect of our method by visualizing attention maps and feature embeddings. In subsection \emph{Shrinking Towards True Behavior}, we briefly discuss the theoretical understanding related with shrinkage estimation. Then we provide more statistical evidences to validate the advantage of the disentangled positive representation. 
%In \ref{sec5.3}, we explain why TAD outperforms baseline methods significantly on some datasets while marginally on others. We find this is closely related with the domain similarity of the source and target dataset.
In subsection \emph{Ablation Study}, we empirically analyze why the disentanglement component is essential. 

\begin{figure}[t]
  \centering
  \subfigure[Stanford Cars]{
    \includegraphics[width=1.5in]{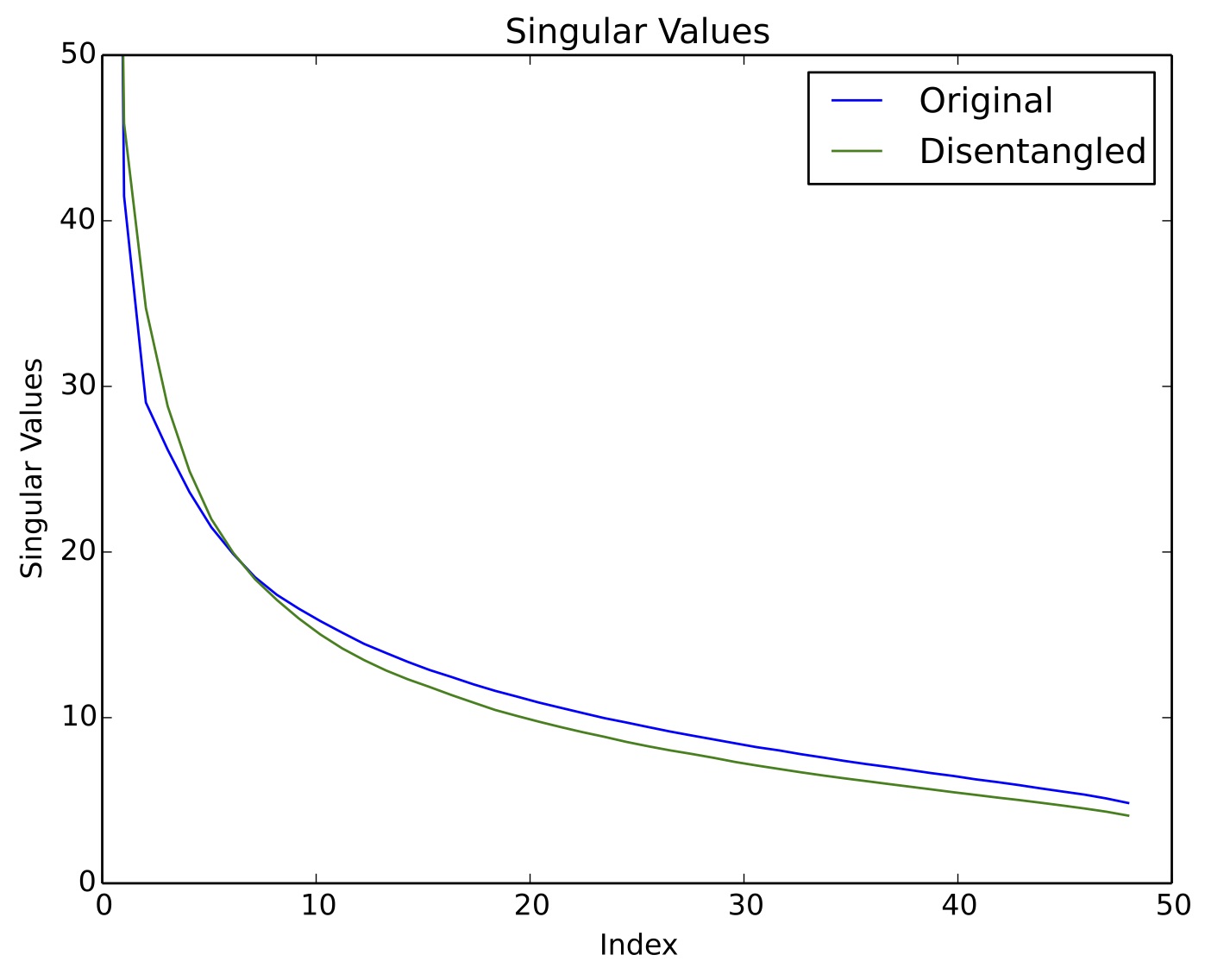}
  }
  \subfigure[CUB-200-2011]{
    \includegraphics[width=1.5in]{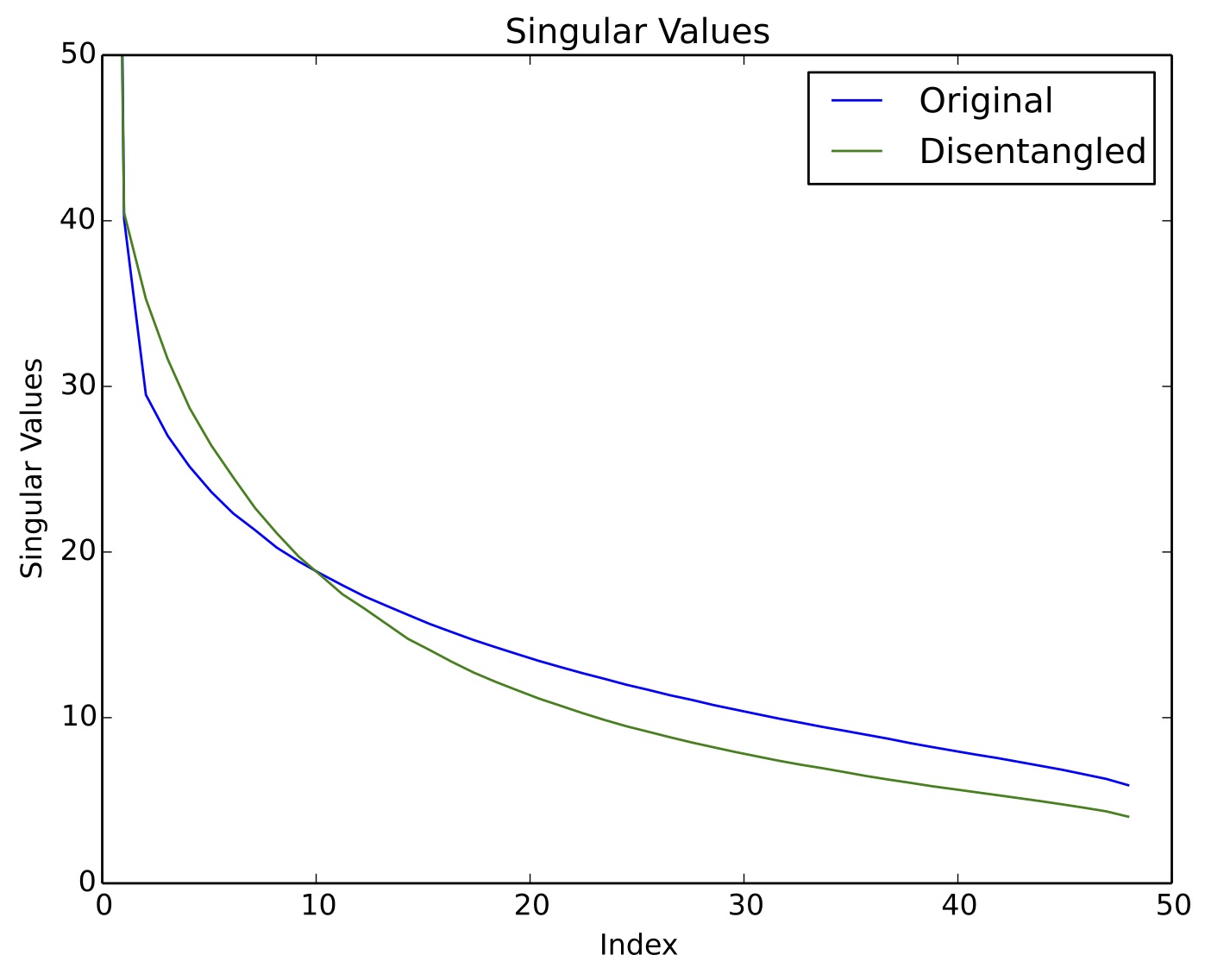}
  }
  \vspace{-2mm}
  \caption{Singular values of the original and disentangled deep representation.}
  \label{fig:svd}
\vspace{-6mm}
\end{figure}

\subsection{Representation Visualization}
\label{sec5.1}
\textbf{Show Cases}. Authors in~\cite{zagoruyko2016paying} show that the spatial attention map plays a critical role in knowledge transfer. We demonstrate the effect of representation disentanglement by visualizing the attention map in Fig~\ref{fig:showcase}. As observed in typical cases from CUB-200-2011 and Stanford Cars, the original representations generated by the ImageNet pre-trained model usually contain a wide range of semantic features, such as objects or backgrounds, in addition to parts of birds. Our proposed disentangler is able to ``purify" the interested concepts into the positive part, while the negative part pays more attention to the complementary constituent. 

\textbf{Embedding Visualization}. Since the most important change of our method is to use the disentangled rather than the original representation as the reference, we are interested in comparing the properties of these two representations on the target task. We visualize the original and disentangled feature representations of Flower-102 and MIT Indoor-67. The dimension of features is reduced along the spatial direction and then plotted in the 2D space using t-SNE embeddings. As illustrated in Figure~\ref{fig:tsne}, deep representations derived by our proposed disentangler are separated more clearly than the original ones for different categories and clustered more tightly for samples of the same category. 

\subsection{Shrinking Towards True Behavior}
\label{sec5.2}
Recent work~\cite{li2018explicit} discusses the connection between their proposed \LTWOSP\ and classical statistical theory of shrinkage estimation~\cite{efron1977stein}. The key hypothesis is that shrinking towards a value which is close to the ``true parameters" is more effective than an arbitrary one. ~\cite{li2018explicit} argues that the starting point is supposed to be more close to the ``true parameters" than zero.~\cite{zagoruyko2016paying,li2019delta}  regularize the feature rather than the parameter, which can be interpreted as shrinking towards the ``true behavior". Our proposed \TheName\ further improves them by explicitly disentangling ``\emph{truer} behavior" by utilizing the global distribution and supervision information of the target dataset. To support the claim, We provide some additional evidences as followed. 

\textbf{Reducing Untransferable Components}. We compute singular eigenvectors and values of the deep representation by SVD. All singular values are sorted in descending order and plotted in Fig~\ref{fig:svd}. Authors in~\cite{chen2019catastrophic} demonstrate that the spectral components corresponding to smaller singular values are less transferable. They find that these less transferable components can be suppressed by involving more training examples. Interestingly, we find similar trends by the proposed representation disentanglement. As observed in Fig~\ref{fig:svd}, smaller singular values of the disentangled positive representation are further reduced compared with the original representation. Fig~\ref{fig:svd} also shows the phenomenon that spectral components corresponding to larger singular values are increased, which does not exist in~\cite{chen2019catastrophic}. This is intuitively consistent to the hypothesis that features relevant to the target task are disentangled and strengthened. 

\begin{figure}[t]
%\centering
\includegraphics[width=0.5\textwidth]{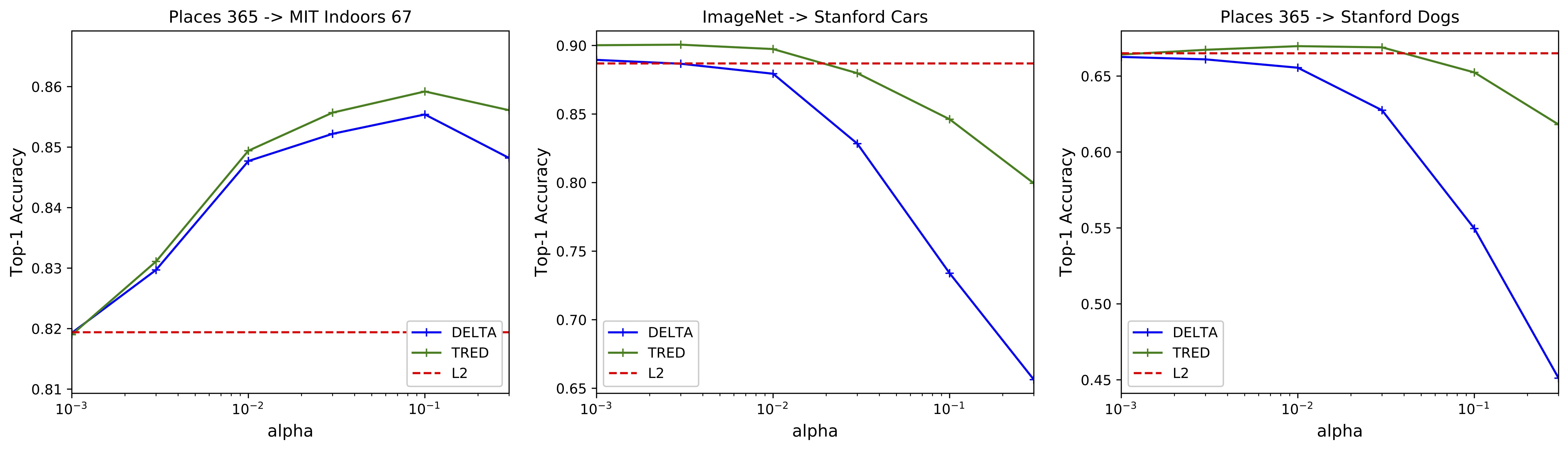}
%\vspace{-5mm}
%\caption{Transfer Learning Performance over Different Regularization Strength $\alpha$.}
\caption{Top-1 accuracy of transfer learning tasks corresponding to different regularization strength $\alpha$.}
\label{fig:negative_transfer}
%\vspace{-4mm}
\end{figure}

\textbf{Robustness to Regularization Strength}. We also provide an empirical evidence to illustrate the effect of ``truer behavior" obtained by our proposed disentangler. The intuition is very straightforward that, if the behavior (representation) used as the reference is ``truer", it is supposed to be more robust to the larger regularization strength. We compare with \DELTA\ which uses the original representation as the reference. We select three transfer learning tasks for evaluation, which are Places365 $\rightarrow$ MIT indoor-67, ImageNet $\rightarrow$ Stanford Cars and Places365 $\rightarrow$ Stanford Dogs. The regularization strength $\alpha$ is gradually increased from 0.001 to 1. As illustrated in Fig~\ref{fig:negative_transfer}, the performance of \DELTA\ falls rapidly as $\alpha$ increases, especially in ImageNet $\rightarrow$ Stanford Cars and Places365 $\rightarrow$ Stanford Dog,  indicating that the regularizer using original representations as the reference suffers from negative transfer seriously. While \TheName\ performs much more robust to the increasing of $\alpha$. %For example, the top-1 accuracy of \TheName\ is about 17\% higher than \DELTA\ in Places365 $\rightarrow$ Stanford Dog at $\alpha=1$.

\iffalse
\begin{figure}[t]
  \centering
  \subfigure[Comparison with \LTWO]{
    \label{fig:ds1}
    \includegraphics[width=1.5in]{res/ds1.pdf}
  }
  \subfigure[Comparison with \LTWOSP/\DELTA]{
    \label{fig:ds2}
    \includegraphics[width=1.5in]{res/ds2.pdf}
  }
  \caption{Relation of improvements brought by \TheName\ and the domain similarity between the source and target dataset.}
  \label{fig:domainsimilarity}
\end{figure}
\fi

\begin{table}
\centering
\caption{Top-1 accuracy (\%) of the target and source task. \TheName- refers to \TheName\ without disentanglement.} 
\begin{threeparttable}
\begin{tabular}{l|c|c|c|c}
\multirow{2}*{Dataset} & \multicolumn{2}{c|}{Target Task} & \multicolumn{2}{c}{Source Task} \\
\cline{2-5}
 & \TheName\ & \TheName- & \TheName\ & \TheName- \\\hline
Flower-102 & 91.34 & 89.79 & 71.23 & 69.73 \\
CUB-200-2011 & 82.07 & 78.84 & 72.65 & 70.11 \\
%FGVC Aircraft & 83.26 & 82.48 & 72.40 & 70.20 \\
MIT Inddors 67\tnote{*} & 85.79 & 82.42 & 28.92 & 25.57
\end{tabular}
 \begin{tablenotes}
        \footnotesize
        \item[*] Pre-trained on Places365 and evaluated on ImageNet.
      \end{tablenotes}
\end{threeparttable}
\label{table:ablation}
\vspace{-3mm}
\end{table}

\subsection{Mechanism Analysis}
\label{sec5.4}
In this part, we briefly discuss about the necessity and relationship of the main components in our method. As our purpose is to disentangle the knowledge which is relevant to the target task, the supervision from the target dataset is of course necessary. We will first discuss whether a simple supervision is enough to ``disentangle'' the relevant knowledge from the source model. Next we will explain why the component of reconstruction is essential to ensure the effectiveness.
 
\textbf{Why disentanglement is useful}. 
It seems reasonable to obtain the discriminative representation only using the classifier corresponding to $L^D_{ce}$. This is equivalent to perform a pre-adaptation upon the source model before fine-tuning. Unfortunately, such straightforward adaptation in a pure supervised manner is prone to over-fitting the limited target examples, as which the resultant representation is not adequate to serve as the prior knowledge. Encouraging the disentanglement between the relevant and irrelevant part, however, provides a distribution-level guarantee to simultaneously preserve the generalization capacity of the source model and adapt for the new task. That is achieved by restricting the integrity of the underlying data structure in a self-supervised way, e.g. disentanglement and reconstruction. To verify the hypothesis, we conduct an ablation study to compare the simpler framework without the disentanglement part, which performs direct transformation on the original representation. This version is denoted by \TheName-. 

We can observe in Table~\ref{table:ablation} that, all evaluated tasks get significant performance drop on the target task without disentanglement. A reasonable guess is that, disentangling helps preserve knowledge in the source model and restrain the representation transformation from over-fitting the classifier $C$. To verify this hypothesis, we compare the top-1 accuracy of ImageNet classification (the source task) between \TheName\ (we only use \TheName-MMD here) and \TheName-. Specifically, we train a random initialized classifier to recognize the category of ImageNet, using the fixed transformed representation as input. The top-1 accuracy of the pre-trained ResNet-101 model is 75.99\%. As shown in Table~\ref{table:ablation}, \TheName- gets more performance drops on ImageNet than \TheName, indicating that representation disentanglement performs better in preserving the general knowledge learned by the source task. 

\textbf{The interdependence between disentanglement and reconstruction}. Disentanglement and reconstruction compose a complete self-supervised requirement. Without the force of disentanglement between the positive and negative part, the transformation will fall into the case of no disentanglement. Therefore, the reconstruction requirement alone is not capable of ensuring the preservation of the source knowledge of the learned positive part. Removing the reconstruction requirement results in an analogy situation, as the transformation can be arbitrary, pursuing to enlarge the discrepancy or independence between the two parts with no knowledge constraint. Intuitively, given that the positive part focus on the supervision information, the negative part can be easily learned to adapt either one (alone) of the two requirements. That causes both cases are almost equivalent to removing the entire disentangler.  

In condition of the reconstruction guarantee, maximizing the attention discrepancy or minimizing the semantic independence between the positive and negative part encourages the two parts to focus on patterns related with different semantic concepts, with as less overlaps or interactions as possible. This is exact a form of representation disentanglement. For example the positive part is activated on heads and feathers of birds, while the negative part is activated on objects such as trees and wires as shown in Fig.~\ref{fig:showcase}. Moreover, this is achieved at a distribution level so that the semantic separation is consistent in the entire dataset. 
\section{Conclusion}
In this paper, we extend the study of catastrophic forgetting and negative transfer in inductive transfer learning. Specifically, we propose a novel approach \TheName\ to regularize the disentangled deep representation, achieving accurate knowledge transfer. We succeed to implement the target-awareness disentanglement, by maximizing the Maximum Mean Discrepancy (MMD) on visual attentions and minimizing the Mutual Information (MI) on semantic features. Extensive experimental results on various real-world transfer learning datasets show that \TheName\ significantly outperforms the state-of-the-art transfer learning regularizers. Moreover, we provide empirical analysis to verify that the disentangled target-awareness representation is closer to the  expected ``true behavior" of the target task.

%In this paper, we present a novel regularization mechanism \TheName\ that preserves the disentangled deep representation produced by the source model. We design a simple and effective framework to disentangle the positive components from the original representation. \TheName\ simultaneously takes the consideration of preventing from losing knowledge of the source model and avoiding negative transfer. Extensive experiments evaluated \TheName\ using various real-world transfer learning datasets. Results show that \TheName\ significantly outperforms the state-of-the-art transfer learning regularizers.

%\appendices
%\input{sections/9_appendix.tex}

\section*{Acknowledgment}
This work is supported in part by the Science and Technology Development Fund of Macau SAR (File no. 0015/2019/AKP to Chengzhong Xu). Parts of experiments in this paper were carried out on Baidu Data Federation Platform (Baidu FedCube). For usages, please contact us via \{fedcube,shubang\}@baidu.com.

\ifCLASSOPTIONcaptionsoff
  \newpage
\fi



\bibliographystyle{IEEEtran}
\bibliography{IEEEabrv}
%


\begin{IEEEbiography}{Michael Shell}
Biography text here.
\end{IEEEbiography}

\begin{IEEEbiographynophoto}{John Doe}
Biography text here.
\end{IEEEbiographynophoto}

%\newpage

\begin{IEEEbiographynophoto}{Jane Doe}
Biography text here.
\end{IEEEbiographynophoto}


%\vfill


\end{document}